\documentclass[journal]{IEEEtran}

\usepackage{amsmath,amssymb,amsfonts}
\usepackage{graphicx}
\usepackage{booktabs}
\usepackage{hyperref}
\usepackage{microtype}
\usepackage{xurl}
\usepackage[ruled,linesnumbered]{algorithm2e}
\usepackage{subfigure}

\hypersetup{colorlinks=true,linkcolor=blue,citecolor=blue,urlcolor=blue}
\begin{document}

\title{Space Is Intelligence: Neural Semigroup Superposition for Riemannian Metric Generation}

\author{Chenghao~Xu
\thanks{C. Xu is with the School of Artificial Intelligence and Robotics and National Engineering Research Center of Robot Visual Perception and Control Technology, Hunan University, Changsha, Hunan, China. Email: chenghaoxu@hnu.edu.cn.}}

\maketitle

\begin{abstract}
Traditional approaches place intelligence in the agent, whether as a learned policy or a search procedure. We instead place intelligence in the space itself: a scene induces a Riemannian metric on the configuration manifold, and action reduces to following the geodesics of that metric rather than invoking a separate planner or collision checker. A single Encoder-Router network realizes this idea through three complementary parameter groups --- frame parameters that orient the generators, modulation parameters that govern their spatial propagation, and basic coefficients that determine their strength. These groups combine through a shared semigroup-superposition mechanism to produce a single Riemannian metric field, yielding a compact architecture whose geometry scales naturally with scene complexity. Trained on a single two-obstacle scene, the model demonstrates robust zero-shot generalization across unseen obstacle configurations, with orders-of-magnitude separation between collision-free and obstacle-penetrating path costs.
\end{abstract}

\begin{IEEEkeywords}
Riemannian metrics, semigroup superposition, Lie algebra, motion planning, zero-shot generalization, learned metric generation
\end{IEEEkeywords}

\section{Introduction}

A foundational assumption underlies most existing approaches to decision-making and control: intelligence resides in the agent. Whether instantiated as a learned neural policy, a graph-search planner (A*, RRT), or a generative world model such as UniSim \cite{yang2024learning} or Genie \cite{bruce2024genie}, the agent is the seat of ``knowing what to do.'' The scene --- obstacles, goal, constraints --- enters as input to this agent-driven decision process.

We propose a fundamentally different view: \textbf{intelligence resides in the space itself}. The planner reduces to a numerical geodesic solver. Given a scene, the environment \textit{directly induces} a Riemannian metric tensor $G(s)$ on the configuration manifold. The optimal path is then simply a geodesic of $G(s)$ --- naturally avoiding obstacles, steering toward the goal, and respecting the constraints encoded in the metric. The planner does not need to ``know'' where obstacles are; the geometry already contains that information. No explicit collision checking is needed: the metric \textit{is} the obstacle.

The metric field is built through \textbf{semigroup superposition}. Each semantic element of the scene (obstacle, goal) independently generates a Lie algebra element $H_k(s)$. These contributions are summed in the Lie algebra and exponentiated into the group to produce $G(s) = \exp(\sum_k H_k(s))$. Because Lie algebra addition is closed under the vector-space structure, the same rule works for \textit{any} number of scene elements---the compositional guarantee behind zero-shot generalization. Inside each $H_k(s)$, a fixed pool of generator matrices---iso (isotropic expansion), aniso (directional stretch), and shear---acts as an alphabet. The exponential map $\exp: \mathrm{sym}(2) \to \mathrm{SPD}(2)$ guarantees that $G(s)$ is symmetric positive-definite for \textbf{any} network output.

Our architecture realizes this view through a single Encoder-Router network, whose two stages share a common superposition structure. The Encoder parses the scene through per-keypoint Sim(2) reference frames: each obstacle centre and the goal independently learn a continuous transformation (tx, ty, scale, theta) that aligns local coordinates with the scene structure, discovered automatically through gradient descent. The Router then produces parameters for each generator slot in three complementary groups. \textbf{Frame parameters} determine where and in which direction each generator acts: per-slot angular offsets rotate each generator's blended direction, and per-slot range modulators control the spatial extent of its influence. \textbf{Modulation parameters} govern how far each generator propagates by controlling the decay profile of its spatial kernel, determining how sharply the metric barrier rises near an obstacle. \textbf{Basic coefficients}---amplitude, sign, and gate---determine each generator's effective contribution strength. All three groups combine through semigroup superposition into a single output: the Riemannian metric field $G(s)$, which is then read by a passive geodesic solver.

\subsection{Contributions}

\begin{enumerate}
    \item \textbf{Scene-conditioned metric generation.}
    A neural architecture that maps a scene description to a valid Riemannian metric field $G(s)$ via semigroup superposition, enabling geodesic path planning without explicit collision checking.

    \item \textbf{Unified metric construction pipeline.}
    A single Encoder-Router pipeline that produces frame parameters, modulation parameters, and generator coefficients, all of which combine into a valid Riemannian metric $G(s)$ through a shared semigroup-superposition rule.

    \item \textbf{Zero-shot compositional generalization.}
    Zero-shot compositional generalization from a single multi-obstacle training scene, with perfect success on unseen obstacle configurations and orders-of-magnitude separation between collision-free and obstacle-penetrating paths.
\end{enumerate}

\section{Related Work}

\subsection{Agent-Centric Approaches}

\textbf{Classical paradigms.} The dominant approaches to motion planning locate intelligence in the agent. Search-based planners such as A* \cite{hart1968formal} and RRT \cite{lavalle1998rapidly} search over graphs or sampled trees built from the free space; their behaviour is determined by hand-designed cost functions and collision-checking procedures. Potential field methods \cite{khatib1986real} embed obstacles and goals as repulsive and attractive force fields, guiding the agent via gradient descent---geometrically intuitive, but reactive and prone to local minima. Model predictive control (MPC) solves a receding-horizon optimization at each step \cite{mayne2000constrained}, yet the cost function and constraint formulation must be engineered per task. Reinforcement learning \cite{sutton2018reinforcement} trains a neural policy $\pi(a \mid s)$ through environment interaction; the policy itself becomes the carrier of behavioural intelligence. Trajectory optimization methods such as CHOMP \cite{ratliff2009chomp} and STOMP \cite{kalakrishnan2011stomp} optimize smoothness and obstacle-related cost functionals over trajectories---the cost landscape is geometrically informed, but the objective must still be specified analytically for each new task. Across all of these, a shared pattern persists: the scene is an input to an agent-driven decision process, and the agent is where ``knowing what to do'' resides.

\textbf{Cutting-edge learned methods.} Recent advances in neural planning and robot foundation models have dramatically improved generalization and scalability, yet they preserve the same agent-centric structure at a larger scale. Learned neural planners such as MPNet \cite{qureshi2019motion} train a neural network to directly output collision-free paths, replacing explicit search with a learned mapping from scene to trajectory. Diffusion policies \cite{chi2023diffusion} and vision-language-action (VLA) models such as RT-2 \cite{brohan2023rt2}, Octo \cite{octo2023}, and $\pi_0$ \cite{black2024pi0} are prominent recent examples of large-scale robot learning systems that map observations to actions. Yet in every case, the architecture is a mapping from observation to action---a function $f: \text{scene} \to \text{action}$ whose parameters carry the system's decision-making capability. The scene determines what to do; the agent determines how.

\textbf{Works closest to ours.} Several lines of work bring geometric or metric reasoning into planning and control. In most cases, however, the geometric object plays an auxiliary role: it enhances the agent's performance without replacing the agent as the locus of intelligence. Metric learning methods such as LMNN \cite{weinberger2009distance} learn a global Mahalanobis distance matrix, which is then used by a downstream classifier---the classifier, not the metric, makes the decision. Inverse reinforcement learning \cite{ziebart2008maximum} recovers a scalar cost function from demonstrations, but the recovered cost is used to train or derive a policy; the policy remains the decision-making entity, and the scalar cost lacks the directional structure of an SPD matrix. Metric learning for text documents \cite{lebanon2006learning} similarly learns task-adapted geometries in feature space for downstream classification and retrieval, rather than using the metric itself as the decision-making mechanism. Equivariant networks \cite{cohen2016group, finzi2020generalizing} encode group structure into neural architectures, but this structure constrains feature representations for a downstream policy or classifier---it does not constitute the decision-making mechanism itself.

Among existing frameworks, Riemannian Motion Policies (RMP)~\cite{ratliff2018riemannian} bears the closest structural resemblance to ours: it also combines Riemannian metrics with motion generation, but does so in the service of closed-loop acceleration policies rather than geometric path planning. An RMP is defined as a pair $(\mathbf{a}, \mathbf{A})$: an acceleration policy $\mathbf{a}(\mathbf{q}, \dot{\mathbf{q}})$ and a Riemannian metric $\mathbf{A}(\mathbf{q}, \dot{\mathbf{q}})$. When multiple RMPs (e.g.\ obstacle avoidance, goal attraction) must be combined, the total acceleration is $\ddot{\mathbf{q}} = (\sum_i \mathbf{A}_i)^{-1} \sum_i \mathbf{A}_i \mathbf{a}_i$: the metric serves as a fusion weight that arbitrates among competing acceleration commands. The fundamental loop is therefore a closed-loop controller---at each timestep the system computes an acceleration, executes it, and recomputes. The metric is an auxiliary object in the service of acceleration policies.

Our framework differs from RMP in two structural dimensions that together reflect a different ontological foundation. \textbf{Core output: space versus action.} RMP's output is an acceleration vector; the metric is a fusion weight. Our output \textit{is} the metric field $G(s)$. There are no acceleration policies, no state-feedback rules, no runtime arbitration. The metric is the sole object produced by the network. The planner merely reads the geometry that the metric defines. \textbf{Extensibility.} Because RMP outputs an action, new requirements must be integrated through additional control structure around that action-generation loop. Because our framework outputs a space, its geometry can in principle be enriched at the level of representation itself rather than only through external control modules.

The contrast is therefore not only technical but architectural. RMP remains control-theoretic at its core: state, action, and acceleration, wrapped in Riemannian geometry. Ours is geometric: the metric is the problem, the geodesic is the solution, and everything else is a passive reader.

Implicit neural representations for planning such as NTFields \cite{ni2023ntfields} also produce a field from which paths are extracted. Their field is a scalar arrival-time field, which cannot encode directional cost structure (e.g.\ normal-direction of an obstacle being more expensive than its tangent). To our knowledge, no prior work places a learned, scene-conditioned, group-structured Riemannian metric as the sole carrier of planning intelligence, with the planner reduced to a passive geodesic solver.

\subsection{Metric as Foundation}

Parallel to the agent-centric tradition runs a mathematical lineage in which the metric \textit{is} the problem. Riemann \cite{riemann1854} established the central role of the metric tensor in determining intrinsic geometry. In the Jacobi-Maupertuis tradition \cite{jacobi1837}, classical trajectories under a potential can be recast in geometric terms, with the potential deforming the effective metric through which motion unfolds. These works do not derive our architecture directly, but they motivate the broader viewpoint that geometry and group structure can serve as primary organizing principles rather than auxiliary decorations.

Our framework synthesizes these two lineages. We inherit from the agent-centric tradition the need to act in realistic, multi-obstacle scenes. We inherit from the mathematical tradition the insight that the metric itself---not a policy, not a scalar cost, not a hand-designed potential---is the fundamental object. What those traditions provide as fixed form, we provide as learned function: the metric field $G(s)$ is the output of a neural network, trained end-to-end to produce metrics whose geodesics solve the task.

\section{Method}

\subsection{Problem Formulation}

Let $\mathcal{M} \subset \mathbb{R}^2$ be the configuration manifold. A \textbf{scene} $S = \{e_1, \ldots, e_K\}$ consists of $K$ semantic elements ($K_o$ obstacles and one goal), each with a position $p_k \in \mathcal{M}$ and radius $r_k$.

The objective is to construct a Riemannian metric $G_S: \mathcal{M} \to \mathrm{SPD}(2)$ such that for any start $s_0$, the geodesic $\gamma^*$ minimizing
\begin{equation}
    \mathcal{L}(\gamma) = \int_0^1 \sqrt{\dot{\gamma}(t)^\top G_S(\gamma(t)) \dot{\gamma}(t)} \, dt
    \label{eq:path_cost}
\end{equation}
is collision-free with respect to all obstacles in $S$.

\subsection{Generator Pool}

The Lie algebra $\mathfrak{gl}(2)$ admits a Cartan decomposition into $\mathrm{sym}(2) \oplus \mathrm{so}(2)$ --- the 3-dimensional symmetric space (stretching) and the 1-dimensional skew-symmetric space (rotation). Our generators live in $\mathrm{sym}(2)$:

\begin{equation}
    B_{\text{iso}} = \begin{bmatrix} 1 & 0 \\ 0 & 1 \end{bmatrix},\;
    B_{\text{aniso}} = \begin{bmatrix} 1 & 0 \\ 0 & -1 \end{bmatrix},\;
    B_{\text{shear}} = \begin{bmatrix} 0 & 1 \\ 1 & 0 \end{bmatrix}
    \label{eq:generators}
\end{equation}

These three matrices form a basis for \textit{all} $2 \times 2$ symmetric matrices. They are \textbf{fixed and non-learnable} --- analogous to an alphabet whose letters do not change, but whose combinations produce infinitely many valid sentences.

\subsection{Semigroup Superposition}

Each of the 9 slots (3 generators $\times$ 3 independent instances) produces one Lie algebra contribution at query point $s$:
\begin{equation}
    H_i(s) = g_i \cdot \alpha_i \cdot \mathrm{sign}_i \cdot K_i(s) \cdot R(\theta_i) \cdot B_{\text{gen}(i)}
    \label{eq:h_slot}
\end{equation}
where $g_i \in [0,1]$ is a learned sigmoid gate, $\alpha_i$ is the amplitude, $\mathrm{sign}_i \in [-1,1]$ controls attraction vs.\ repulsion, $K_i(s)$ is a learned spatial kernel that blends a goal-attraction component and an obstacle-repulsion component via a softmax over two kernel sources, $R(\theta_i)$ rotates the generator into a direction blended from three direction bases (goal-direction, obstacle-normal, obstacle-tangent) through learned softmax weights, and $B_{\text{gen}(i)}$ is the fixed generator matrix (Eq.\,\ref{eq:generators}). Each kernel component follows an exponential-decay profile parameterised by its decay rate, range, exponent shape, and kernel sharpness $\sigma_i$.

The total metric is assembled via Lie algebra addition and the exponential map:
\begin{equation}
    H(s) = \sum_{i=1}^{9} H_i(s), \qquad G(s) = \exp(H(s))
    \label{eq:superposition}
\end{equation}

Crucially, $\exp: \mathrm{sym}(2) \to \mathrm{SPD}(2)$ guarantees $G(s)$ is symmetric positive-definite for \textbf{any} $H(s)$. This is a theorem --- not a regularizer, not a projection. The neural network provides any 3 numbers $(a,b,c)$; the exponential map bakes them into a guaranteed-valid metric.

\subsection{Parameter Structure}

The scene is parsed by a per-keypoint Encoder. Each obstacle centre and the goal independently predict a continuous Sim(2) reference frame, transform scene coordinates into that frame, and pass the result through a shared MLP to produce a 16-dimensional embedding. A single Router --- a shared MLP body $16 \to 64 \to 64$ feeding 11 parallel output heads --- processes each keypoint embedding and outputs parameters in three groups:

\vspace{4pt}
\noindent\textbf{Frame parameters.}
The Encoder's per-keypoint Sim(2) reference frame $(\mathrm{tx}_k, \mathrm{ty}_k, \mathrm{scale}_k, \mathrm{theta}_k)$ determines how the scene is perceived from each keypoint's perspective, positioning and orienting the local coordinate system before encoding. Independently, the Router outputs per-slot angular offsets $\Delta\theta_i$ and range modulators $\Delta s_i$ that refine each generator's direction and spatial reach. In the Composer, each generator's blended direction is rotated by $\Delta\theta_i$ and its spatial range is scaled by $\Delta s_i$ before the kernel is evaluated. Together, these two levels of frame parameters determine where each generator acts and in which direction it exerts force.

\vspace{4pt}
\noindent\textbf{Modulation parameters.}
Per-slot kernel sharpness $\sigma_i$ controls the decay of each generator's spatial kernel. Larger $\sigma_i$ produces steeper falloff, concentrating the generator's effect around obstacles and creating sharper metric barriers.

\vspace{4pt}
\noindent\textbf{Basic coefficients.}
Amplitude $\alpha_i$, sign $\mathrm{sign}_i \in [-1, 1]$ (attraction vs.\ repulsion), and sigmoid gate $g_i \in [0, 1]$ jointly determine the generator's effective contribution strength. The gate provides a learnable slot-selection mechanism: a variance penalty on gate values discourages collapse while preserving gradient competition among active slots.

\vspace{4pt}
All three groups are produced by the same Router from the same Encoder, and a single computation yields the full parameter set:
\begin{equation}
    \{\text{frame}, \text{modulation}, \text{coefficients}\} \leftarrow \text{Router}(\text{Encoder}(S)).
\end{equation}
Each keypoint independently outputs a full parameter set. The Composer then assembles the metric field $G(s)$ point by point. At each query position $s$ and for each slot $i$, it evaluates the spatial kernel $K_i(s)$ using the slot's modulation parameters, rotates the fixed generator $B_{\text{gen}(i)}$ into the blended direction determined by the frame parameters, scales the result by the basic coefficients, and sums all contributions in the Lie algebra. The sum $H(s) = \sum_i H_i(s)$ is exponentiated via $\exp$ to produce the SPD matrix $G(s)$. The metric field is the sole output of the system; it is then read by a passive geodesic solver.

\subsection{Training}

We train with a contrastive loss on positive (collision-free) and negative (obstacle-penetrating) path pairs:
\begin{equation}
    \mathcal{L} = \frac{\overline{\mathcal{L}}^+}{\overline{\mathcal{L}}^- + \epsilon}, \qquad \epsilon = 0.1
    \label{eq:loss}
\end{equation}
The total loss is $\mathcal{L}_{\text{total}} = \mathcal{L} + \lambda \cdot \mathcal{L}_{\text{bal}}$, where the gate-balance regularizer $\mathcal{L}_{\text{bal}} = \sum_i (g_i - \bar{g})^2$ with $\lambda = 0.01$ discourages gate collapse and maintains active slot competition. We use the Adam optimizer with learning rate $3\times10^{-3}$ and gradient clipping at 10.0. Training data are generated from a single scene: 24 positive paths sampled via A*, 48 negative paths from random search, and 101 discretised waypoints per path with 0.05 additive noise. A three-stage staircase curriculum of 200 epochs per stage progressively unlocks direction bases --- goal-direction $\to$ obstacle-normal $\to$ obstacle-tangent --- with the Encoder and Router backbone frozen from stage 2 onward. The total training budget is 600 epochs on \textbf{one} scene with 2 obstacles; all evaluations are zero-shot.

\begin{algorithm}[t]
\caption{Forward Pass: Scene $\to$ Metric Field}
\label{alg:forward}
\KwIn{Scene $S = \{(\mathbf{p}_k, r_k)_{k=1}^{K_o}, \mathbf{p}_{\text{goal}}\}$, query $\mathbf{s}$}
\KwOut{Riemannian metric $G(\mathbf{s}) \in \mathrm{SPD}(2)$}
\BlankLine

\tcp{Encoder: per-keypoint Sim(2) frames}
\For{each keypoint $k \in \{1..K_o,\ \text{goal}\}$}{
    $(\text{tx}, \text{ty}, s, \theta)_k \gets \textsc{RefParamHead}(\mathbf{f}_k)$\;
    $\tilde{\mathbf{f}}_k \gets \textsc{Sim2Trans}(\mathbf{f}_k;\,\text{tx},\text{ty},s,\theta)$\;
    $\mathbf{e}_k \gets \textsc{MLP}_{128\to128\to16}(\tilde{\mathbf{f}}_k)$\label{alg:enc}\;
}

\tcp{Router: shared body $\to$ multi-head output}
$\mathbf{h} \gets \textsc{MLP}_{16\to64\to64}(\mathbf{e}_k)$\label{alg:rtr}
\For{each slot $i = 1..9$}{
    $\alpha_i, d_i, r_i \gets \text{softplus}(\cdot) + \varepsilon$\;
    $g_i \gets \sigma(\cdot),\; \text{sgn}_i \gets \tanh(\cdot)$\;
    $\mathbf{w}^\text{dir}_i \gets \text{softmax}(\cdot)$\;
    $\mathbf{w}^\text{ker}_i \gets \text{softmax}(\cdot)$\;
    $p_i \gets \sigma(\cdot)\cdot 4.5 + 0.5$\;
    $\sigma_i \gets \sigma(\cdot)\cdot 9.9 + 0.1$\;
    $\Delta\theta_i \gets \tanh(\cdot) \cdot \frac{\pi}{6}$\;
    $\Delta s_i \gets \exp(\tanh(\cdot) \cdot \ln 2)$\;
}

\tcp{Composer: Lie algebra accumulation}
$\mathbf{H} \gets \mathbf{0}_{2\times2}$\label{alg:cmp}\;
$\mathbf{v}_\text{goal}, \mathbf{v}_\text{norm}, \mathbf{v}_\text{tang} \gets$ directions at $\mathbf{s}$\;
\For{each slot $i = 1..9$}{
    $\mathbf{v}_i \gets \text{norm}(\sum_{b} w^\text{dir}_{i,b}\, \mathbf{v}_b)$\;
    $\mathbf{v}_i \gets \textsc{Rot}(\mathbf{v}_i,\Delta\theta_i)$\;
    $\hat{r}_i \gets r_i \cdot \Delta s_i$\;
    $K_i \gets \sum_{b} w^\text{ker}_{i,b}\, \exp\!\big(-\tilde{\sigma}_{i,b}\, d_i (\frac{d^2_{\text{src}_b}}{\tilde{r}_{i,b}^2})^{p_i/2}\big)$\;
    $\mathbf{R}_i \gets \textsc{RotSym}(B_{\text{gen}(i)},\,\mathbf{v}_i)$\;
    $\mathbf{H} \gets \mathbf{H} + g_i\,\alpha_i\,\text{sgn}_i\,K_i\,\mathbf{R}_i$\;
}
$G(\mathbf{s}) \gets \textsc{ExpSym}_{2\times2}(\mathbf{H})$\;
\Return $G(\mathbf{s})$\;\label{alg:end}
\end{algorithm}

Algorithm~\ref{alg:forward} provides the complete forward pass for reference. The algorithm proceeds in three stages. \textbf{Encoding} (lines~\ref{alg:enc}--\ref{alg:rtr}): each keypoint independently predicts a continuous Sim(2) reference frame $(\text{tx},\text{ty},s,\theta)_k$ via a small MLP, transforms the raw scene features into that keypoint's local coordinate system, and passes them through a shared MLP ($128\to128\to16$) to produce a 16-dimensional embedding $\mathbf{e}_k$. \textbf{Routing} (lines~\ref{alg:rtr}--\ref{alg:cmp}): a shared 2-layer MLP ($16\to64\to64$) maps each embedding to a hidden state $\mathbf{h}$, from which 11 parallel linear heads extract per-slot parameters --- basic coefficients (amplitude $\alpha_i$, gate $g_i$, sign $\text{sgn}_i$, decay $d_i$, range $r_i$), modulation parameters (kernel exponent $p_i \in [0.5,5.0]$, sharpness $\sigma_i \in [0.1,10.0]$), frame parameters (angular offset $\Delta\theta_i \in [-\frac{\pi}{6},\frac{\pi}{6}]$, range modulator $\Delta s_i \in [0.5,2.0]$), and softmax mixing weights for the three direction bases and two kernel sources. 

Then, \textbf{Composition} (lines~\ref{alg:cmp}--\ref{alg:end}): at each query point $\mathbf{s}$, the three direction bases (goal-direction, obstacle-normal, obstacle-tangent) are blended per slot via the learned softmax weights, rotated by $\Delta\theta_i$, and used to evaluate the spatial kernel as a learned blend of goal and obstacle kernels. The goal kernel is parameterised by $\hat{r}_i = r_i \Delta s_i$; the obstacle kernel uses the obstacle's geometric radius as its spatial scale and applies a per-keypoint, spatially decaying sharpness modulation. The fixed generator $B_{\text{gen}(i)}$ is rotated into the blended direction, scaled by the effective contribution $g_i \alpha_i \text{sgn}_i K_i$, and accumulated in the Lie algebra $\mathbf{H}$. Finally, $\textsc{ExpSym}_{2\times2}$, a closed-form $2\times2$ symmetric matrix exponential, maps $\mathbf{H}$ to a guaranteed SPD metric $G(\mathbf{s})$.

\section{Experiments}

\begin{figure*}[t]
\centering
\includegraphics[width=\textwidth]{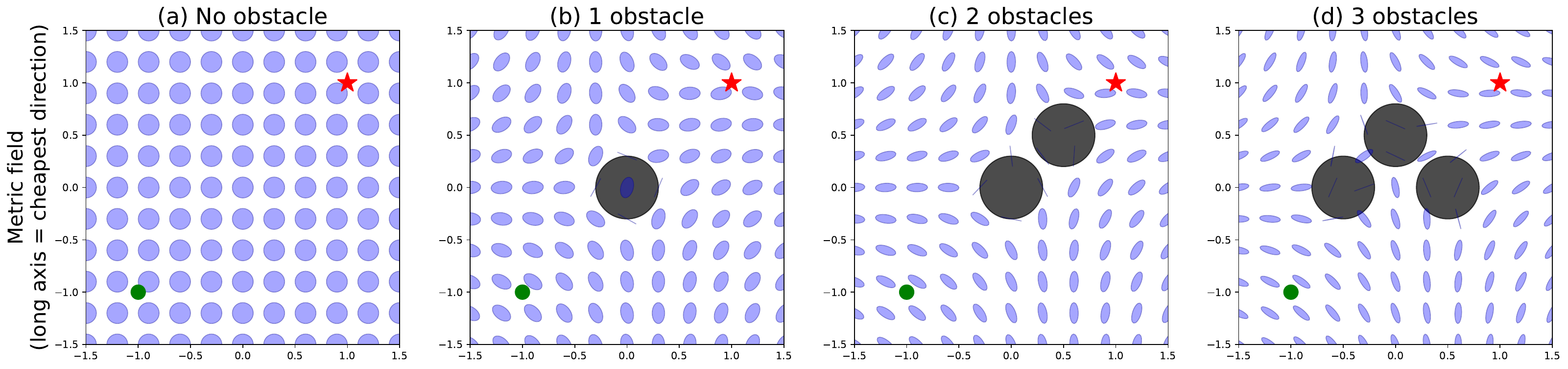}
\caption{\textbf{Learned metric field across four scenes.}
Blue ellipses = local SPD(2) metric $G(s)$: the long axis is the cheapest direction to move, the short axis the most expensive, and the degree of elongation reflects the directional anisotropy.}
\label{fig:metric_mod}
\end{figure*}

\subsection{Setup}

We evaluate on a 2D planar point robot in a $3 \times 3$ world. The training scene has 2 obstacles at $(0, 0)$ and $(0.5, 0.5)$, start $(-1, -1)$, and goal $(1, 1)$. Twelve zero-shot test scenes are designed to systematically cover structural variations: 1 scene with 0 obstacles (goal only), 4 scenes with a single obstacle at different positions (centre, shifted, corner, near-boundary), 4 scenes with two obstacles arranged in distinct spatial patterns (aligned, diagonal, vertical, and an unseen combination), and 2 scenes with three obstacles (clustered and spread apart). This progression tests whether the model generalises across obstacle count, position, density, and spatial arrangement --- all of which vary substantially across the test set without retraining.

Our model uses 9 slots (3 generators $\times$ 3 independent instances), 16-dim embeddings, a per-keypoint Encoder with shared MLP (128 $\to$ 128 $\to$ 16), and a Router with shared body ($16 \to 64 \to 64$) feeding 11 parallel output heads. A* path planning on a $101 \times 101$ grid evaluates actual geodesic distances under the learned metric.

\subsection{Ablation: Sparse Gating}

As a control experiment, we tested sparse top-$k$ gating, which restricts each scene to exactly $k=6$ active slots via hard sparsemax, hypothesizing that explicit slot selection would improve interpretability. Our standard dense gate allows all 9 slots ($k=9$, one per generator instance) to be active simultaneously via independent sigmoid gates. Under the same training protocol, the sparse top-6 variant succeeds on only 10 of 12 scenes, whereas the dense baseline succeeds on all 12. The root cause is that hard sparsity permanently freezes the parameters of masked-out slots (21 parameters, zero gradient forever), transforming a continuous optimization over the full Lie algebra into a discrete selection problem with no gradient competition. The design lesson is clear: \textit{let all slots compete through continuous gradients --- not human prior --- to decide which generators contribute to the metric.}

\subsection{Metric Geometry and Modulation}

Figure~\ref{fig:metric_mod} visualises the learned geometry for four scenes (0, 1, 2, 3 obstacles). In free regions, metric ellipses are near-circular, forming a smooth gradient toward the goal. Near obstacles, ellipses become elongated tangentially --- the short axis points toward the obstacle (making traversal expensive), while the long axis aligns along the tangent (making circumvention cheap) --- pushing paths to go \textit{around} rather than through.

\subsection{Zero-Shot Generalization}

\begin{figure}[t]
\centering
\includegraphics[width=\columnwidth]{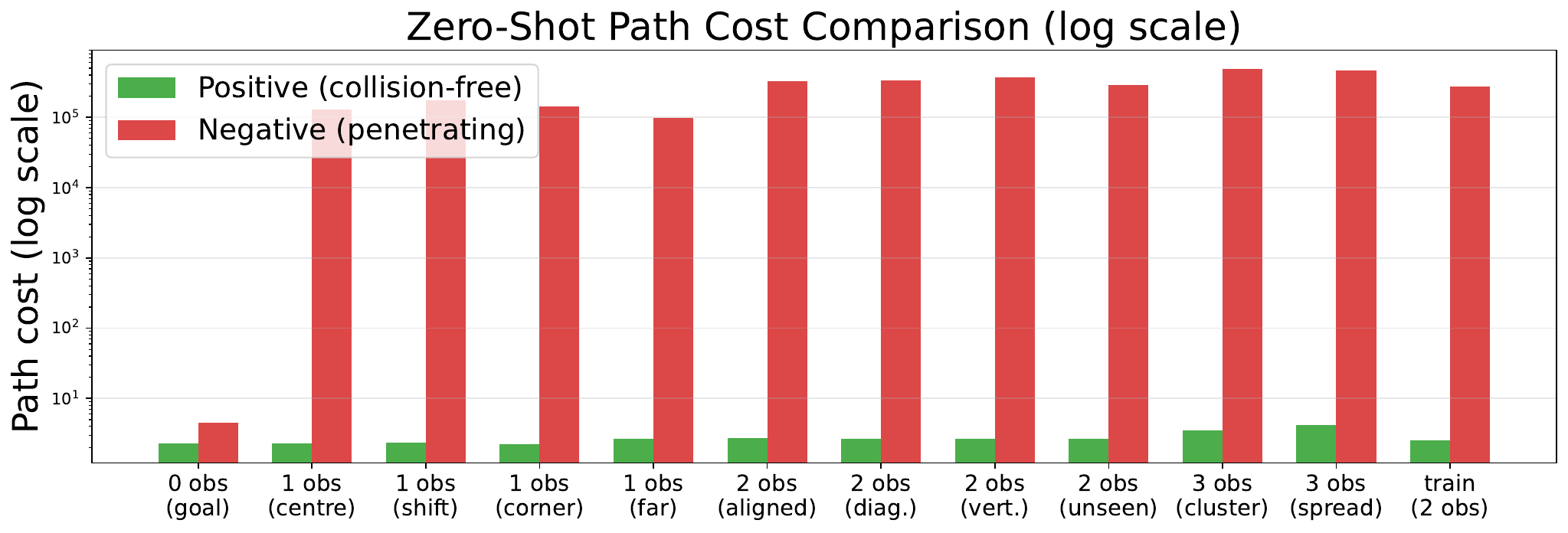}
\caption{\textbf{Zero-shot path cost comparison across 12 scenes.}
Positive (green, collision-free) and negative (red, obstacle-penetrating) path costs on log scale.}
\label{fig:gen}
\end{figure}

\begin{table}[t]
\centering
\caption{Per-scene breakdown. All 12 scenes succeed.}
\label{tab:batch}
\begin{tabular}{@{}lrrr@{}}
\toprule
Scene & Obstacles & pos cost & neg cost \\
\midrule
Goal only     & 0 & 2.3 & $4.5\times 10^0$ \\
1 obs (centre)& 1 & 2.3 & $1.3\times 10^5$ \\
1 obs (shift) & 1 & 2.4 & $1.8\times 10^5$ \\
1 obs (corner)& 1 & 2.2 & $1.4\times 10^5$ \\
1 obs (far)   & 1 & 2.6 & $9.7\times 10^4$ \\
2 obs (aligned) & 2 & 2.7 & $3.3\times 10^5$ \\
2 obs (diag.)  & 2 & 2.7 & $3.4\times 10^5$ \\
2 obs (vert.)  & 2 & 2.7 & $3.7\times 10^5$ \\
2 obs (unseen) & 2 & 2.7 & $2.9\times 10^5$ \\
3 obs (cluster)& 3 & 3.5 & $4.9\times 10^5$ \\
3 obs (spread) & 3 & 4.2 & $4.7\times 10^5$ \\
Train (2 obs)  & 2 & 2.6 & $2.7\times 10^5$ \\
\bottomrule
\end{tabular}
\end{table}

Figure~\ref{fig:gen} and Table~\ref{tab:batch} present the quantitative evaluation. Positive (collision-free) path costs range from $2.3$ to $4.2$; negative (obstacle-penetrating) path costs range from $4.9\times 10^5$ (3-obstacle cluster scene) down to $9.7\times 10^4$ (1-obstacle far scene) in scenes that actually contain obstacles, yielding a cost separation of 3--5 orders of magnitude. The goal-only scene is the single exception: with no obstacles present, the ``negative'' path cannot genuinely penetrate anything, so its cost ($4.5$) naturally collapses to near-Euclidean. For any scene with at least one obstacle, the worst positive cost ($4.2$, in the 3-obstacle spread scene) remains separated from the best negative cost ($9.7\times 10^4$) by a factor greater than $2\times 10^4$. The training scene alone achieves $2.6$ vs.\ $2.7\times 10^5$, a separation exceeding $10^5\times$. This margin means that an A* planner relying on the learned metric can safely discriminate collision-free paths from penetrating ones across all obstacle-containing scenes without threshold tuning --- the metric itself provides the boundary. All 12 scenes succeed, including the zero-obstacle case, where the metric decays to near-Euclidean with gentle goal attraction.

This confirms that the semigroup-superposition rule $G(s)=\exp(\sum_k H_k(s))$ correctly handles varying numbers of scene elements without retraining.

\subsection{Geodesic Paths}

\begin{figure*}[t]
\centering
\includegraphics[width=\textwidth]{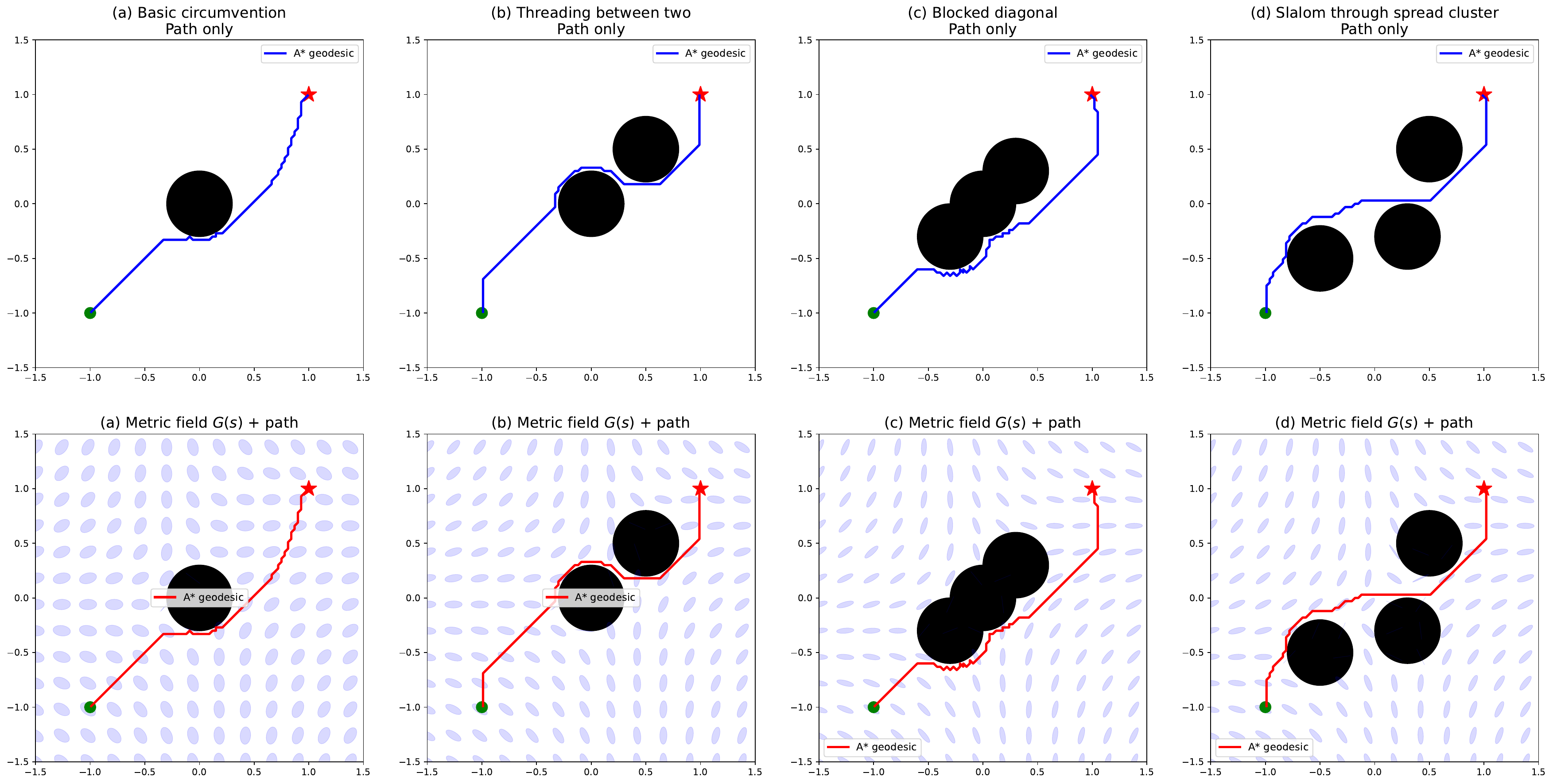}
\caption{\textbf{Optimal paths under the learned metric across four scenes.}
Top row: A* geodesic path only.
Bottom row: path overlaid on metric ellipses.
(a) 1 obstacle --- basic circumvention.
(b) 2 obstacles --- threading between two diagonally placed obstacles.
(c) 3 obstacles blocking the start--goal diagonal, forcing a wide detour.
(d) 3 obstacles spread across the scene --- the geodesic slaloms between them, weaving through the narrowest corridor.}
\label{fig:paths}
\end{figure*}

Figure~\ref{fig:paths} extracts A*-optimal geodesics under $G(s)$ for four qualitatively distinct scenes. The metric field itself encodes collision avoidance: near obstacles, ellipses stretch tangentially, turning direct traversal into an expensive direction and pushing the geodesic outward. In (a), a single obstacle at the origin is smoothly circumvented. In (b), the path threads between two diagonal obstacles. In (c), three obstacles placed along the diagonal create a solid barrier on the shortest path; the geodesic detours widely around the blockade. In (d), three obstacles spread across the scene create competing repulsive fields; the geodesic slaloms through the narrowest corridor, weaving side-to-side between them. In all cases, collision avoidance is an emergent property of the metric geometry, not an explicit planning step.

\subsection{Training Dynamics and Learned Parameters}

\begin{figure}[t]
\centering
\includegraphics[width=\columnwidth]{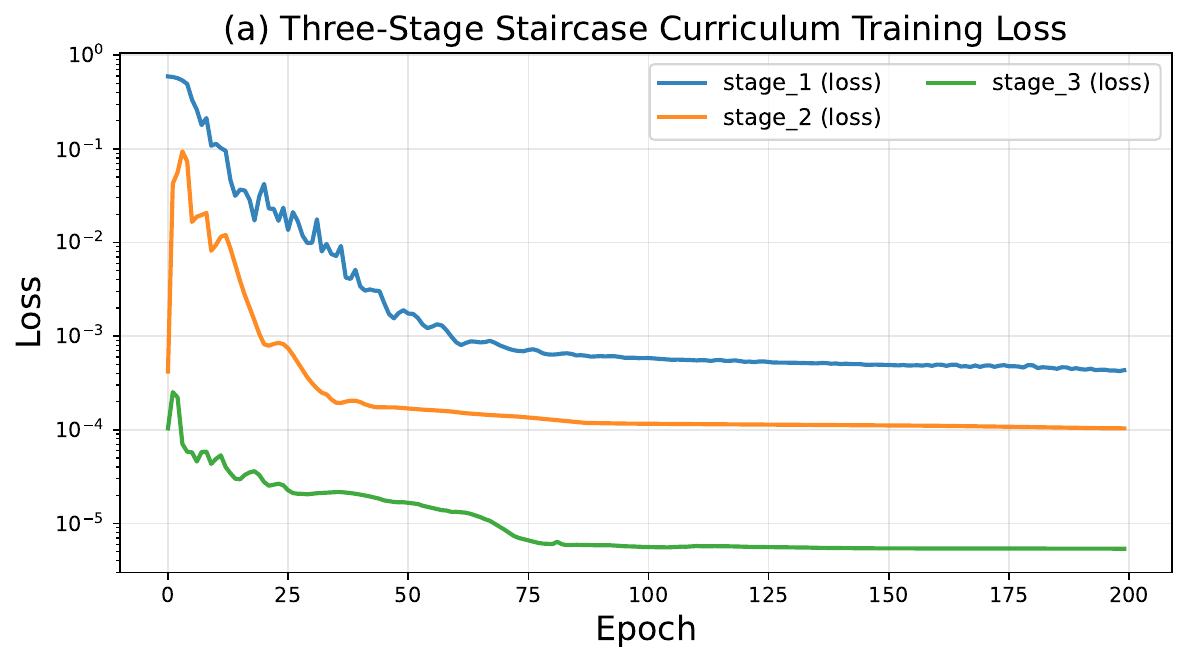}
\caption{\textbf{Three-stage staircase curriculum training loss (log scale).}
Stage 1 (goal-direction only) converges rapidly. Stages 2 and 3 (adding obstacle-normal and obstacle-tangent) maintain the converged state while refining directional resolution.}
\label{fig:loss}
\end{figure}

\begin{figure}[t]
\centering
\subfigure[Per-slot angular offset $\Delta\theta_i$.]{
    \includegraphics[width=0.8\columnwidth]{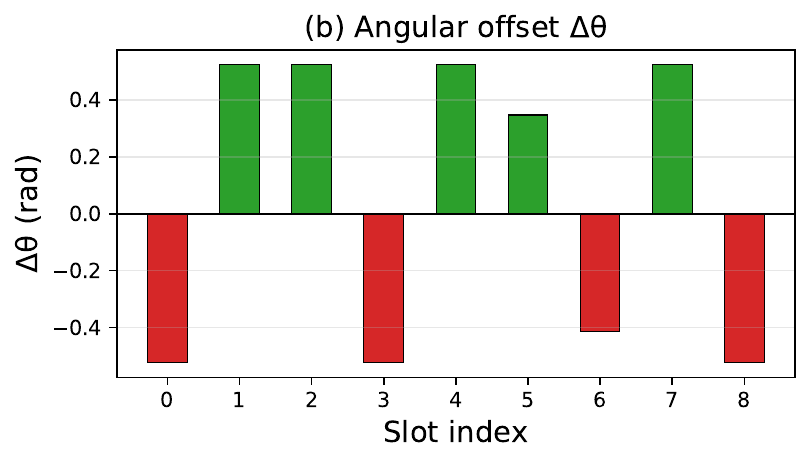}
    \label{fig:frame_dtheta}
}
\subfigure[Per-slot range modulator $\Delta s_i$.]{
    \includegraphics[width=0.8\columnwidth]{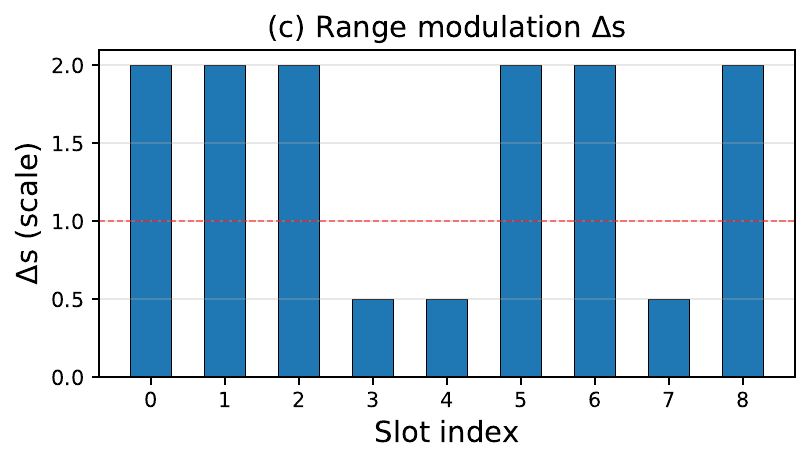}
    \label{fig:frame_dscale}
}
\caption{\textbf{Per-slot frame parameters learned by the Router on the training scene.}
\textbf{(a)} Slots $\{0,3,6\}$ group at $+0.52$~rad, slots $\{1,4,7\}$ at $-0.52$~rad, and slots $\{2,5,8\}$ at intermediate values.
\textbf{(b)} Six slots lie at $2.0\times$ (broad repulsion), while three lie at $0.5\times$ (sharp repulsion near the obstacle surface).}
\label{fig:frames}
\end{figure}

Figure~\ref{fig:loss} shows the staircase curriculum. Stage 1 (goal-direction only) converges rapidly, establishing a global attraction toward the goal. Stage 2 adds obstacle-normal repulsion, and Stage 3 adds obstacle-tangent directional refinement; both maintain the previously converged state, with loss remaining flat across stage boundaries.

Figure~\ref{fig:frames} displays the per-slot frame parameters learned on the training scene. The 9 slots form three functional groups by angular offset $\Delta\theta_i$: slots $\{0,3,6\}$ push repulsion in one tangential direction ($+0.52$~rad), slots $\{1,4,7\}$ push in the opposite direction ($-0.52$~rad), and slots $\{2,5,8\}$ occupy intermediate angles. The range modulators $\Delta s_i$ complement this: six slots expand their kernel range to $2.0\times$ (broad repulsion far from obstacles), while three slots contract to $0.5\times$ (sharp repulsion close to the obstacle surface). Together, the angular offsets and range modulators give the 9 slots complementary directional and spatial specialisation --- each slot focuses on a different angular sector and distance regime around each obstacle. Without these frame parameters, the blended direction of each generator leaks repulsion into the free corridor, inflating positive path costs by 5--17$\times$ (see Section~\ref{sec:frame_ablation}).

\subsection{Ablation: Frame Parameters}
\label{sec:frame_ablation}

To isolate the contribution of the frame parameters, we trained a control model with the frame heads frozen at identity ($\Delta\theta_i = 0$, $\Delta s_i = 1.0$), matched on all other hyperparameters. Table~\ref{tab:ablation_frame} reports the results.

\begin{table}[t]
\centering
\caption{Frame parameter ablation: with vs.\ without per-slot frame parameters. Both configurations use the same architecture and training protocol.}
\label{tab:ablation_frame}
\begin{tabular}{@{}lcccc@{}}
\toprule
& \multicolumn{2}{c}{Full model (with frame)} & \multicolumn{2}{c}{Ablation (no frame)} \\
\cmidrule(lr){2-3} \cmidrule(lr){4-5}
Scene & pos $\times$ & neg $\times$ & pos $\times$ & neg $\times$ \\
\midrule
1 obstacle & 2.3 & $1.3\times 10^5$ & 12.7 & $1.1\times 10^7$ \\
2 obstacles & 2.7 & $3.3\times 10^5$ & 38.6 & $1.1\times 10^8$ \\
3 obstacles & 3.5 & $4.9\times 10^5$ & 46.2 & $5.5\times 10^7$ \\
Goal only & 2.3 & 4.5 & 2.3 & 4.5 \\
Train scene & 2.6 & $2.7\times 10^5$ & 38.9 & $3.4\times 10^7$ \\
\bottomrule
\end{tabular}
\end{table}

Removing the frame parameters increases free-path costs by a factor of 5--17$\times$ across scenes with obstacles. The effect is concentrated on the positive path cost: negative (penetrating) paths remain sufficiently penalised in both conditions, but the frame parameters are essential for keeping free-space channels open. Without angular offsets, the blended direction of each generator leaks repulsion into the free corridor; without range modulation, the generator's influence spreads too broadly. The per-slot frame parameters allow the network to assign specialised roles---some slots focus repulsion sharply near obstacles in the normal direction, others widen the attraction toward the goal---which the ablation demonstrates is not achievable through the basic coefficients alone.

\subsection{Compositional Stress Tests}

To test the compositional guarantee $G(s) = \exp(\sum_k H_k(s))$ beyond the 12 standard scenes, we evaluated our model on denser configurations (4--6 obstacles) and overlapping obstacle pairs (centre-to-centre distances of 0.25 and 0.5 units, with obstacle radius 0.3 producing physical overlap). Table~\ref{tab:stress} summarises the results.

\begin{table}[t]
\centering
\caption{Compositional stress tests: dense obstacles and overlapping obstacle pairs.}
\label{tab:stress}
\begin{tabular}{@{}lccc@{}}
\toprule
Scene & Obstacles & pos cost & neg cost \\
\midrule
Dense 4 (box)   & 4 & 25.7  & $6.5\times 10^5$ \\
Dense 4 (row)   & 4 & 21.4  & $7.5\times 10^5$ \\
Dense 5 (cross) & 5 & 1684  & $7.7\times 10^5$ \\
Dense 5 (mixed) & 5 & 1452  & $7.8\times 10^5$ \\
Dense 6         & 6 & 3100  & $9.4\times 10^5$ \\
Dense 6 (grid)  & 6 & 4385  & $6.6\times 10^5$ \\
Overlap (sep 0.50) & 2 & 2.6   & $4.2\times 10^5$ \\
Overlap (sep 0.25) & 2 & 2.5   & $4.0\times 10^5$ \\
\bottomrule
\end{tabular}
\end{table}

All configurations succeed (6/6 dense scenes and 2/2 overlapping pairs). The metric maintains positive-negative separation above three orders of magnitude even at 6 obstacles, although free-path costs increase with obstacle density as metric barriers from neighbouring obstacles compress the available corridor. In the overlapping pairs, the two obstacles physically intersect (radius 0.3 each, centre separation 0.25--0.50), yet the metric still produces clean free paths with costs of 2.5--2.6 and separation above $10^5\times$, demonstrating that the superposition rule handles spatially overlapping field contributions without special-case logic.

\begin{figure}[t]
\centering
\subfigure[Dense obstacles (4--6).]{
    \includegraphics[width=0.8\columnwidth]{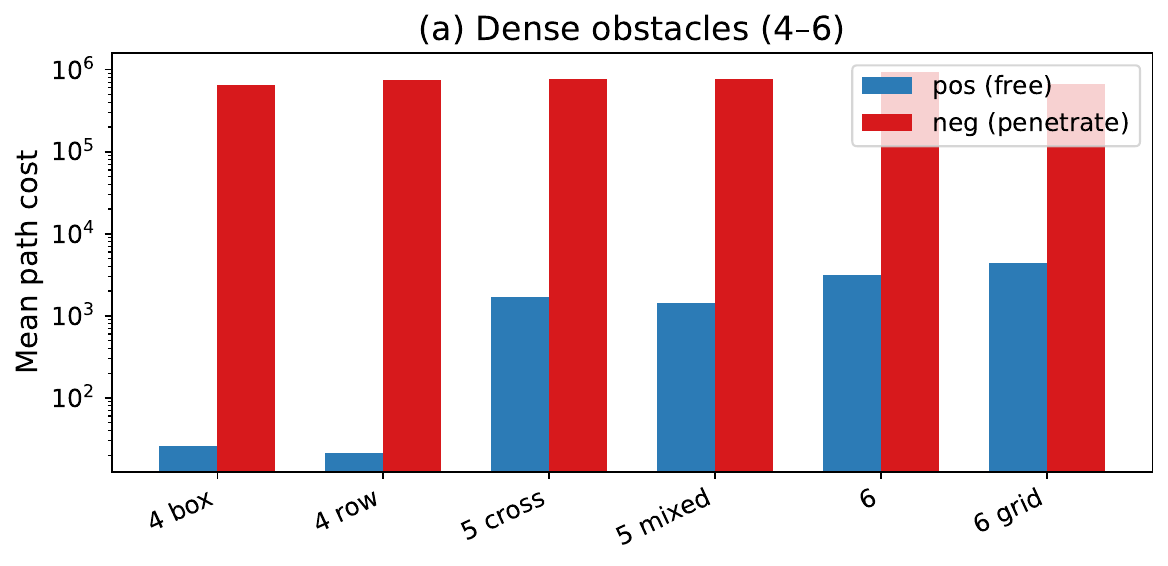}
    \label{fig:dense}
}
\subfigure[Overlapping obstacle pairs (sep 0.25--0.50).]{
    \includegraphics[width=0.8\columnwidth]{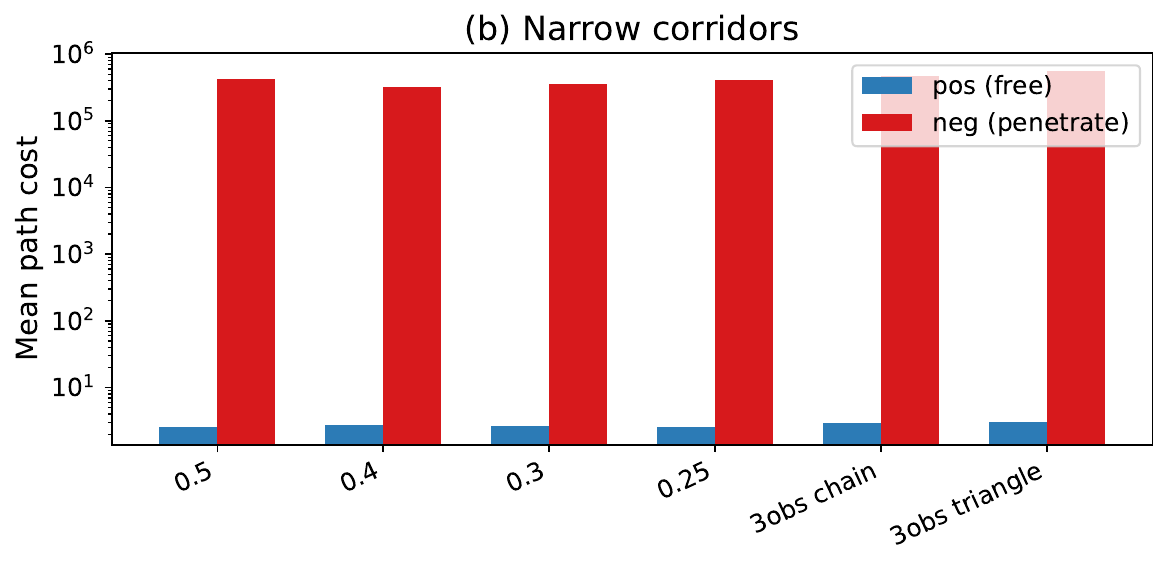}
    \label{fig:narrow}
}
\caption{\textbf{Compositional stress tests.}
\textbf{(a)} Dense obstacle evaluation (4--6 obstacles, log scale).
\textbf{(b)} Overlapping obstacle pair evaluation (centre separation 0.25--0.50).}
\label{fig:stress}
\end{figure}

\subsection{Random Scene Statistics}

To verify that the 12/12 result is not an artefact of hand-selected evaluation scenes, we tested 100 randomly generated scenes (1--6 obstacles with random positions in $[-0.8, 0.8]^2$). The model achieved \textbf{100/100 success} (100\%). The median free-path cost is 3.6, and the geometric mean separation between positive and negative paths exceeds $10^4\times$ across all 100 scenes.

\begin{figure}[t]
\centering
\subfigure[Path cost vs.\ number of obstacles.]{
    \includegraphics[width=0.8\columnwidth]{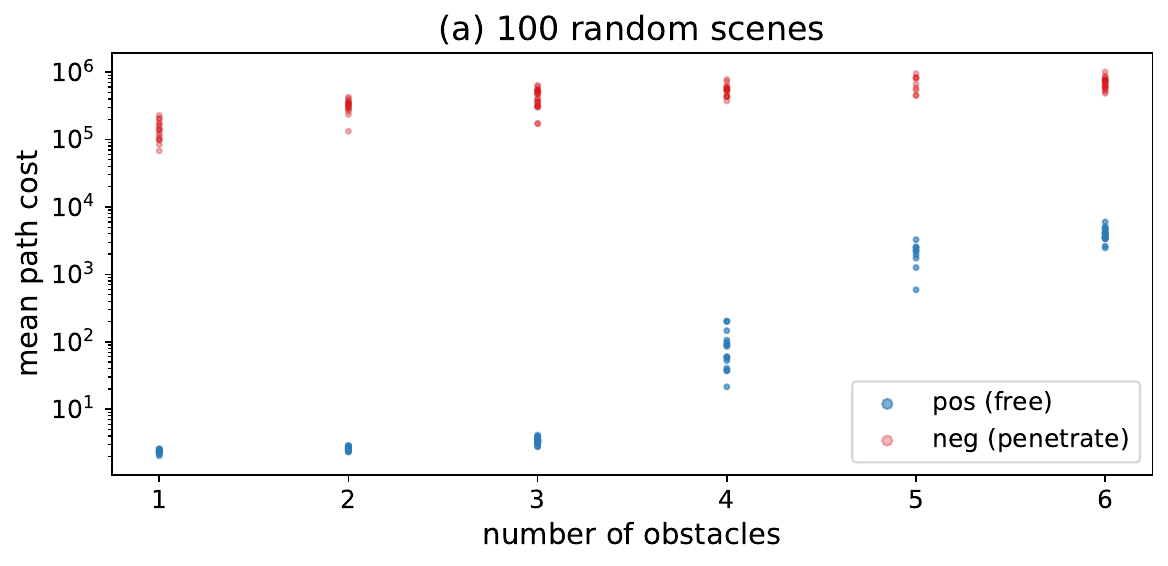}
    \label{fig:random_scatter}
}
\subfigure[Distribution of free-path (pos) costs.]{
    \includegraphics[width=0.8\columnwidth]{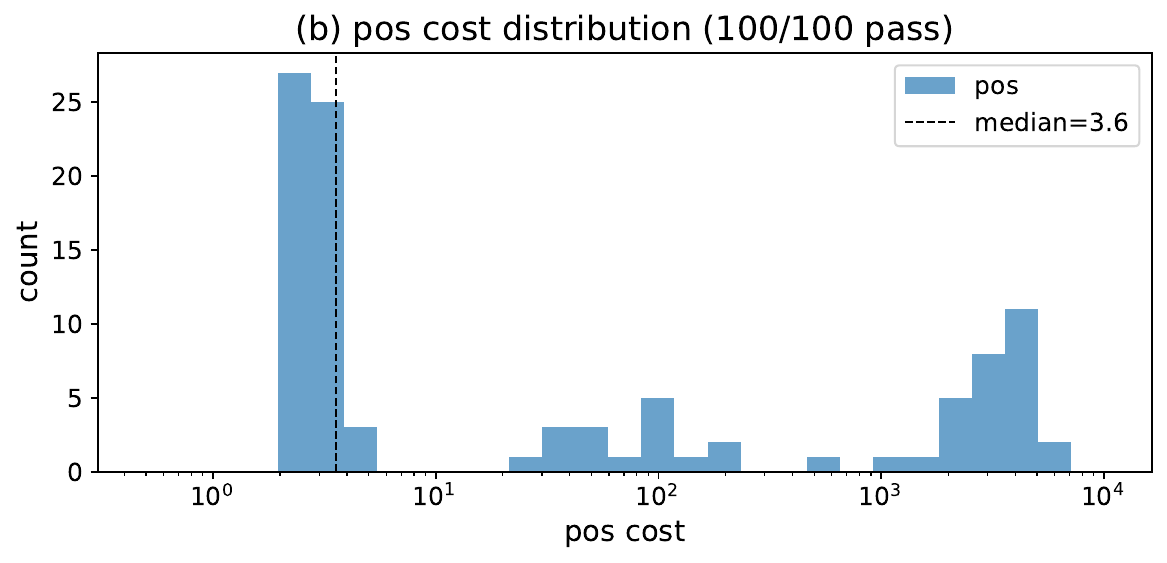}
    \label{fig:random_hist}
}
\caption{\textbf{Random scene statistics across 100 scenes.}
\textbf{(a)} Path cost vs.\ number of obstacles (log scale).
\textbf{(b)} Distribution of free-path costs (100/100 success, median 3.6).}
\label{fig:random}
\end{figure}

\section{Discussion}

\subsection{Why Does It Generalize?}

Zero-shot generalization arises from two structural guarantees. \textbf{Vector space closure.} The Lie algebra $\mathrm{sym}(2)$ is a vector space. For \textit{any} number of scene elements $K$, the sum $H(s) = \sum_{k=1}^K H_k(s)$ is always a valid Lie algebra element. No max, min, or product combination offers the same linear compositional rule. \textbf{Exponential validity.} The exponential map sends every symmetric matrix to a positive-definite one. Once the scene contributions have been summed into $H(s)$, the map $G(s)=\exp(H(s))$ guarantees that the resulting metric is valid everywhere in the configuration space. Compositionality therefore does not come at the cost of feasibility.

\subsection{A Third Paradigm: Geometry as Motion Intelligence}

Motion planning has been dominated by two paradigms: reinforcement learning, which learns a policy $\pi(a \mid s)$ mapping states to actions through trial and error; and model predictive control, which predicts future states through a dynamics model and optimizes a cost function \cite{mayne2000constrained}. Both separate the planning machinery from the world representation.

This framework proposes a third paradigm: \textbf{rather than learning a policy or predicting future states, learn the Riemannian geometry of the scene}. The metric field $G(s)$ \textit{is} the motion intelligence---it directly encodes, at every spatial point, which directions are cheap to move in and which are expensive. Following a geodesic under this metric is the full motion plan; no separate policy evaluation, value iteration, or trajectory optimization is required. The geometric field replaces the planner entirely.

This inverts the conventional approach: intelligence resides no longer in the agent but in the geometry of the space itself. The task is not to solve a planning problem but to discover the geometry hidden in the scene---and once discovered, motion is simply geodesic flow.

\subsection{Generators as Grammar}

The 3 generators function analogously to a linguistic alphabet. The same generator matrix can appear in multiple slots with independent learned parameters (gate, amplitude, direction, etc.), enabling the Router to ``write sentences'' with repeating letters in different contexts. This differs from mixture-of-experts architectures, where experts are selected as modules rather than recombined as shared algebraic primitives. The combinatorial expressivity obtained from 3 generators through 9 independent slots far exceeds the 3-dimensional Lie algebra span alone --- the redundancy is not waste, but the pressure that lets gradient competition discover effective compositions.

\subsection{Limitations}

The current validation is restricted to 2D planar point robots with explicit keypoint positions. Generalization to higher-dimensional configuration spaces and learned perception frontends are important directions for future work.

\section{Conclusion}

We have presented a framework in which a scene induces a Riemannian metric field through neural semigroup superposition. Three fixed Lie algebra generators, composed by an Encoder-Router architecture through 9 independently parameterised slots and three complementary parameter groups (frame, modulation, and basic coefficients), combine into a single metric field whose geodesics produce collision-free motion. Trained on just one two-obstacle scene, the model achieves broad generalization across unseen obstacle configurations, showing that the learned geometry composes robustly beyond the training setup.

The exponential map $\exp$ provides the mathematical guarantee that all outputs are valid metrics, while additive composition in the Lie algebra gives the framework its compositional structure. In this sense, the framework can be viewed as a generative counterpart to Felix Klein's Erlangen Program \cite{klein1893comparative}. Klein asked how a transformation group determines a geometry through its invariants. We ask the complementary question: given a scene, what combination of fixed generators should be composed to produce the geometry required for action? In our model, the Router selects how the generators combine, yielding a scene-conditioned element $G(s) \in \mathrm{SPD}(2)$ that determines the local geometry at each point. Lie algebras, exponential maps, and group actions provide the common mathematical language connecting these two perspectives.

\bibliographystyle{IEEEtran}
\bibliography{references}

\end{document}